%% file: main.tex
\documentclass{article}
\input{preamble}
\usepackage{microtype}
\usepackage{graphicx}
\usepackage{subfigure}
\usepackage{booktabs} % for professional tables
\usepackage{makecell}
\usepackage{hyperref}

\usepackage[accepted]{icml2025}

\usepackage{amsmath}
\usepackage{amssymb}
\usepackage{mathtools}
\usepackage{amsthm}
\usepackage{bm}
\usepackage{graphicx}
\usepackage{textcomp}
\usepackage{xcolor}
\usepackage{wrapfig}
\usepackage{booktabs}
\usepackage{lipsum}
\usepackage{balance}
\usepackage{url}
\usepackage{enumerate}
\usepackage{soul}
\usepackage{enumitem}
\usepackage{verbatimbox}
\usepackage{tcolorbox}
\usepackage{fancyvrb}
\usepackage{fancyref}
\usepackage{hyperref}
\usepackage{subcaption}
\usepackage{amsmath}
\usepackage{amssymb}
\usepackage{amsfonts}
\usepackage{tabularray}
\usepackage{array}
\usepackage[longtable]{multirow}
\usepackage{ragged2e}
\usepackage{longtable}
\usepackage{booktabs}
\usepackage{caption}
\usepackage{multirow}
\usepackage{colortbl}
\usepackage{xparse}
\usepackage[normalem]{ulem}
\usepackage{algpseudocode}
\usepackage{marvosym}
\usepackage{amssymb}
\usepackage[acronym]{glossaries}
\usepackage{pifont}
\usepackage{graphicx}

\usepackage[capitalize,noabbrev]{cleveref}

\theoremstyle{plain}

\theoremstyle{definition}

\theoremstyle{remark}

\usepackage[textsize=tiny]{todonotes}

\icmltitlerunning{}

\begin{document}

\twocolumn[
\icmltitle{Humanoid-VLA: Towards Universal Humanoid Control with Visual Integration}

% It is OKAY to include author information, even for blind
% submissions: the style file will automatically remove it for you
% unless you've provided the [accepted] option to the icml2025
% package.

% List of affiliations: The first argument should be a (short)
% identifier you will use later to specify author affiliations
% Academic affiliations should list Department, University, City, Region, Country
% Industry affiliations should list Company, City, Region, Country

% You can specify symbols, otherwise they are numbered in order.
% Ideally, you should not use this facility. Affiliations will be numbered
% in order of appearance and this is the preferred way.
\icmlsetsymbol{equal}{*}

\begin{icmlauthorlist}
\icmlauthor{Pengxiang Ding}{equal,yyy,sch}
\icmlauthor{Jianfei Ma}{equal,comp}
\icmlauthor{Xinyang Tong}{equal,yyy}
\icmlauthor{Binghong Zou}{comp}
\icmlauthor{Xinxin Luo}{comp}
\icmlauthor{Yiguo Fan}{yyy}
\icmlauthor{Ting Wang}{yyy}
%\icmlauthor{}{sch}
\icmlauthor{Hongchao Lu}{yyy}
\icmlauthor{Panzhong Mo}{comp}
\icmlauthor{Jinxin Liu}{comp}
\icmlauthor{Yuefan Wang}{yyy,sch}
\icmlauthor{Huaicheng Zhou}{comp}
\icmlauthor{Wenshuo Feng}{comp}
\icmlauthor{Jiacheng Liu}{yyy,sch}
\icmlauthor{Siteng Huang}{yyy}
\icmlauthor{Donglin Wang}{yyy,comp}
%\icmlauthor{}{sch}
%\icmlauthor{}{sch}
\end{icmlauthorlist}

\icmlaffiliation{yyy}{Milab, Westlake University}
\icmlaffiliation{comp}{Westlake Robotics}
\icmlaffiliation{sch}{Zhejiang University}

% \icmlcorrespondingauthor{Firstname1 Lastname1}{first1.last1@xxx.edu}
\icmlcorrespondingauthor{Donglin Wang}{wangdonglin@westlake.edu.cn}

% You may provide any keywords that you
% find helpful for describing your paper; these are used to populate
% the "keywords" metadata in the PDF but will not be shown in the document
\icmlkeywords{Machine Learning, ICML}

\vskip 0.3in
]

% this must go after the closing bracket ] following \twocolumn[ ...

% This command actually creates the footnote in the first column
% listing the affiliations and the copyright notice.
% The command takes one argument, which is text to display at the start of the footnote.
% The \icmlEqualContribution command is standard text for equal contribution.
% Remove it (just {}) if you do not need this facility.

% \printAffiliationsAndNotice{}  % leave blank if no need to mention equal contribution
\printAffiliationsAndNotice{\icmlEqualContribution} % otherwise use the standard text.

\input{includes/0_abstract}
\input{includes/1_intro}

\input{includes/2_rw}

\input{includes/3_0_overview}
\input{includes/4_method}

\input{includes/5_exps}

\input{includes/6_conclus}

\section*{Impact Statement}
This paper presents work whose goal is to advance the field of Machine Learning in Robotics. There are many potential societal consequences of our work, none which we feel must be specifically highlighted here.

% \clearpage

% \section*{Impact Statement}
% This paper presents work whose goal is to advance the field of Machine Learning. There are many potential societal consequences of our work, none which we feel must be specifically highlighted here.

\bibliography{example_paper}
\bibliographystyle{icml2025}

\input{includes/8_appendix}

\end{document}

%% file: preamble.tex
\usepackage[T1]{fontenc} 
\newcommand{\modelname}{Humanoid-VLA }

%% file: includes/0_abstract.tex
\begin{abstract}

This paper addresses the limitations of current humanoid robot control frameworks, which primarily rely on reactive mechanisms and lack autonomous interaction capabilities due to data scarcity. 
We propose Humanoid-VLA, a novel framework that integrates language understanding, egocentric scene perception, and motion control, enabling universal humanoid control. 
Humanoid-VLA begins with language-motion pre-alignment using non-egocentric human motion datasets paired with textual descriptions, allowing the model to learn universal motion patterns and action semantics. We then incorporate egocentric visual context through a parameter efficient video-conditioned fine-tuning, enabling context-aware motion generation. Furthermore, we introduce a self-supervised data augmentation strategy that automatically generates pseudo-annotations directly derived from motion data. This process converts raw motion sequences into informative question-answer pairs, facilitating the effective use of large-scale unlabeled video data. 
Built upon whole-body control architectures, extensive experiments show that Humanoid-VLA achieves object interaction and environment exploration tasks with enhanced contextual awareness, demonstrating a more human-like capacity for adaptive and intelligent engagement.

\end{abstract}

%% file: includes/1_intro.tex
\section{Introduction}

\begin{figure}
  \centering
   \includegraphics[width=1\linewidth]{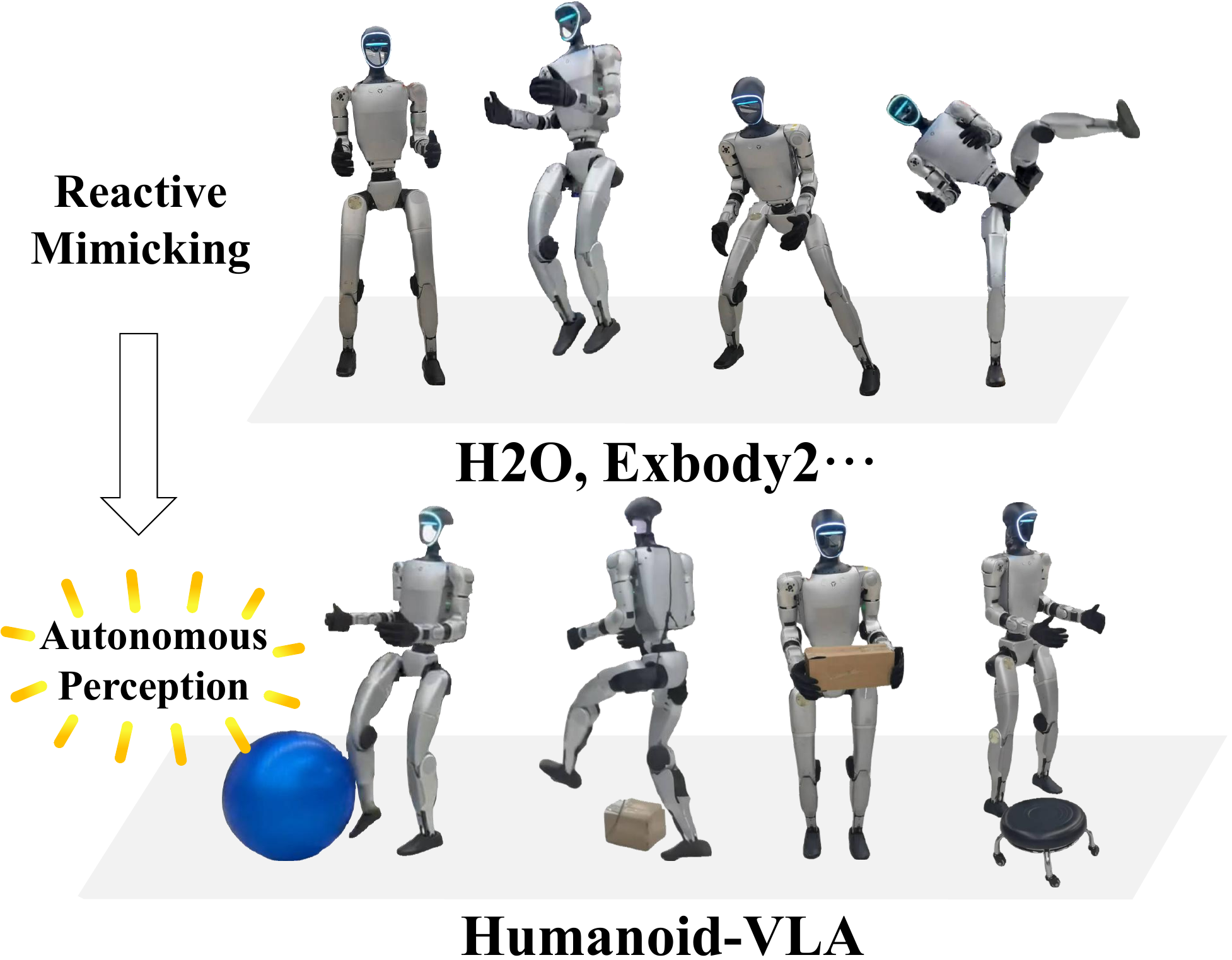}
   \caption{\textbf{Comparison between previous works and our approach.} With the capability of autonomous perception, Humanoid-VLA can perform tasks to interact with objects, significantly advancing beyond previous methods that rely on mimicking human demonstrations for motion execution.}
   \label{fig:teaser}
\end{figure}

Humanoid robots are poised to transform diverse industries—from healthcare to manufacturing—by combining human-like dexterity with adaptability to execute complex tasks. Building on extensive human motion datasets from computer graphics research \cite{mahmood2019amass,guo2020action2motion}, recent advances have established data-driven frameworks for humanoid motion skill acquisition. 

Initial research \cite{cheng2024expressive,ji2024exbody2} developed whole-body controllers that translate basic human kinematic sequences into humanoid motion. The field has since progressed to integrate multimodal perception, enabling humanoids to perform real-time mimicry of human demonstrations \cite{he2024learning} and respond fluidly to natural language commands \cite{mao2024learning}. 
While these approaches achieve high-fidelity motion control in humanoid robots, they operate primarily through reactive mechanisms, dynamically adjusting motions in response to external inputs. They cannot perceive autonomously and infer potential interaction targets within their immediate surroundings. This limitation substantially impedes their deployment in scenarios that demand object manipulation or adaptive exploration in complex environments.
To this end, this paper aims to investigate universal humanoid control with egocentric visual integration. 

However, developing such a system faces a significant bottleneck: \textbf{data scarcity}.
Existing motion capture datasets lack synchronized first-person visual information, making direct transfer to egocentric tasks impossible. Moreover, while teleoperation offers a theoretical pathway for collecting visuomotor data, its prohibitive costs severely constrain large-scale acquisition. These constraints lead to inadequate training datasets in quantity and diversity, which hinders the development of a foundation model for humanoid control with egocentric visual integration. 

As to the challenge of data scarcity, we propose a feasible and cost-effective paradigm.
Specifically, we begin by establishing a language-motion pre-alignment using non-egocentric human motion datasets paired with textual descriptions. This enables the model to learn universal motion patterns and action semantics from diverse third-person observations, yielding a robust, generalizable motion representation that does not rely on egocentric visual input.
Next, we incorporate egocentric visual context through a parameter-efficient cross-attention module. This adaptive mechanism preserves the integrity of the pretrained model while allowing for the dynamic fusion of first-person visual features, thereby enabling context-aware motion generation. 
Our framework essentially reduces the dependence on egocentric datasets, making combining language understanding and egocentric scene perception with motion control feasible.

Furthermore, the existing training paradigm alone is insufficient to ensure optimal model performance, primarily due to limitations in the alignment between motion and language. Drawing inspiration from the success of MLLMs \cite{liu2023llava, zhang2023video}—where robust large language models (LLMs) serve as foundational components—we argue that the effectiveness of our framework hinges on the pre-alignment of motion and language representations. Achieving this alignment, however, heavily depends on the availability of large-scale and high-quality data. Unfortunately, current motion datasets are insufficient in scale to meet this need. While video sources offer a vast reservoir of human data, their utility for model training is constrained by the lack of motion description annotations.

To address this limitation, we propose a self-supervised data augmentation framework that generates pseudo-annotations through automated motion analysis.
Our solution features an automatic annotation pipeline that extracts semantic meaning directly from motion sequences via carefully designed self-supervised tasks.
A representative implementation involves temporarily masking specific body joints within motion sequences and training the model to reconstruct the occluded movements. We automatically generate instructional prompts for such tasks as "missing left arm <Occlusion> motion data. Please complete the motion" paired with corresponding ground truth motions as target outputs. This automated process systematically converts raw motion data into meaningful question-answer pairs.
By integrating these self-supervised learning objectives, our approach circumvents the need for manually annotated textual descriptions while effectively utilizing large-scale and unlabeled motion data extracted from video repositories.

Finally, integrating a whole-body controller following previous work~\cite{he2024learning}, our proposed \textbf{Humanoid-VLA} seamlessly combines language understanding, scene perception, and motion control into a unified system. 
Extensive experiments show our approach significantly enhances humanoid robots' autonomous interaction capabilities in real-world environments, paving the way for practical deployment across diverse applications.

%% file: includes/2_rw.tex
\section{Related Works}

% \begin{figure*}
%   \centering
%    \includegraphics[width=1.0\linewidth]{imgs/data.pdf}
%    \vspace{-5mm}
%    \caption{\textbf{Train}. }
%    \label{fig:}
%    \vspace{-3mm}
% \end{figure*}

% V2
\subsection{Humanoid Control}
Traditional humanoid control methods \cite{li2023dynamic, kuindersma2016optimization, elobaid2023online, dantec2021whole, dai2014whole} like MPC provide accuracy and stability but lack adaptability, while learning-based methods offer flexibility but rely on human motion data due to limited humanoid datasets.
Works like Exbody \citep{cheng2024expressive}, Exbody2 \citep{ji2024exbody2}, HARMON \citep{jiang2024harmon}, and mobile-television \citep{lu2024mobile} perform upper-body motion retargeting for humanoid robots using the SMPL model \citep{loper2023smpl} and root velocity tracking for lower-body locomotion.
To achieve flexible and complex motions, methods such as PHC \citep{luo2023perpetual}, H2O \citep{he2024learning}, and OmniH2O \citep{he2024omnih2o} use the SMPL model to extend motion retargeting to whole-body control.
Additionally, approaches like OmniH2O \citep{he2024omnih2o}, HARMON \citep{jiang2024harmon}, and UH-1 \citep{mao2024learning} enable language-guided motion generation. However, these methods are reactivate, meaning that the models generate various motions passively based on text or key points. 
To perform more advanced tasks autonomously in dynamic and complex environments, \textbf{egocentric visual information} is indispensable.

\subsection{Humanoid Dataset}
Apart from the first-person visual information, aligning motion with semantically relevant textual information is crucial for the construction of a foundational humanoid robot model.
Previous human datasets, such as AMASS \citep{mahmood2019amass},  HumanML3D \citep{Guo_2022_CVPR}, Motion-X \citep{lin2023motion}, and Human3.6M \citep{h36m_pami, IonescuSminchisescu11} provide large-scale human motion data. While some works use human motion retargeting to develop humanoid robot datasets \citep{he2024omnih2o, he2024learning, cheng2024expressive, ji2024exbody2}, these datasets often suffer from sparse text annotations and limited scale, restricting their use in training foundational models. 
Even though some methods can mitigate this issue, they generally suffer from high costs~\cite{mao2024learning} and a lack of precision~\cite{tevet2023human}.
In contrast, we propose a self-suprvised data augmentation method that circumvents the need for manually
annotated textual descriptions while effectively utilizing
large-scale, unlabeled motion data extracted from video repositories for the training of robot foundation model.

\subsection{VLA for Robotics Learning}

In recent years, VLA models have advanced robot learning, particularly for robotic arms and quadrupeds, by integrating vision, language, and action to enhance task and environment generalization. For robotic arms, models like RT-2~\cite{brohan2023rt}, OpenVLA~\cite{kim2024openvla}, GR-2~\cite{cheang2024gr}, RoboMamba~\cite{liurobomamba}, and RDT-1B~\cite{liu2024rdt} leverage visual and language inputs for efficient task execution. In quadrupeds, models such as QUAR-VLA~\cite{ding2025quar} and QUART-Online~\cite{tong2024quart} improve generalization and adaptability in dynamic environments, while $\pi_0$~\cite{black2024pi_0} enables multi-embodied robots to perform diverse tasks. Despite these advances, due to the scarcity of datasets that combine first-person visual information, textual motion descriptions, and whole-body motion data for humanoid robots, VLA models have yet to be applied to humanoid robots. This paper takes the first step in building the Humanoid-VLA model to enable humanoid robots to autonomously perform loco-manipulation tasks.

%% file: includes/4_method.tex
\section{Humanoid-VLA}

\subsection{Preliminary}

\textbf{Definition of humanoid control. }
With the growing availability of human data in the graphics community, recent humanoid control has increasingly adopted methods that learn from human data. 
Specifically, given a target body pose from physical teleoperation (e.g., a motion capture system) and the humanoid's proprioception, the whole-body controller $\mathcal{P}$ generates joint torques to control the humanoid robot. Formally, this can be expressed as
\begin{equation}
    j_t = \mathcal{P}(s_t, p_t),
\end{equation}
where $s_t, p_t, j_t$ means target body
pose, humanoid's proprioception and joint torques at time $t\in \mathbb{N}^+$. 
% p_t dim 242

\textbf{Limitation of humanoid control. } 

However, developing a general-purpose robot requires purposive learning, which involves extracting meaningful intentions from human data and adapting prior experiences to novel tasks or environments. 
Current data acquisition methods, focusing mainly on human joint poses, lack integration with egocentric vision. Thus, they can only teach robots what actions are performed, not the underlying intent or context. Consequently, pose-level imitation is inherently limited in generalizability due to environmental discrepancies.

\begin{figure*}[ht]
  \centering
   \includegraphics[width=0.97\linewidth]{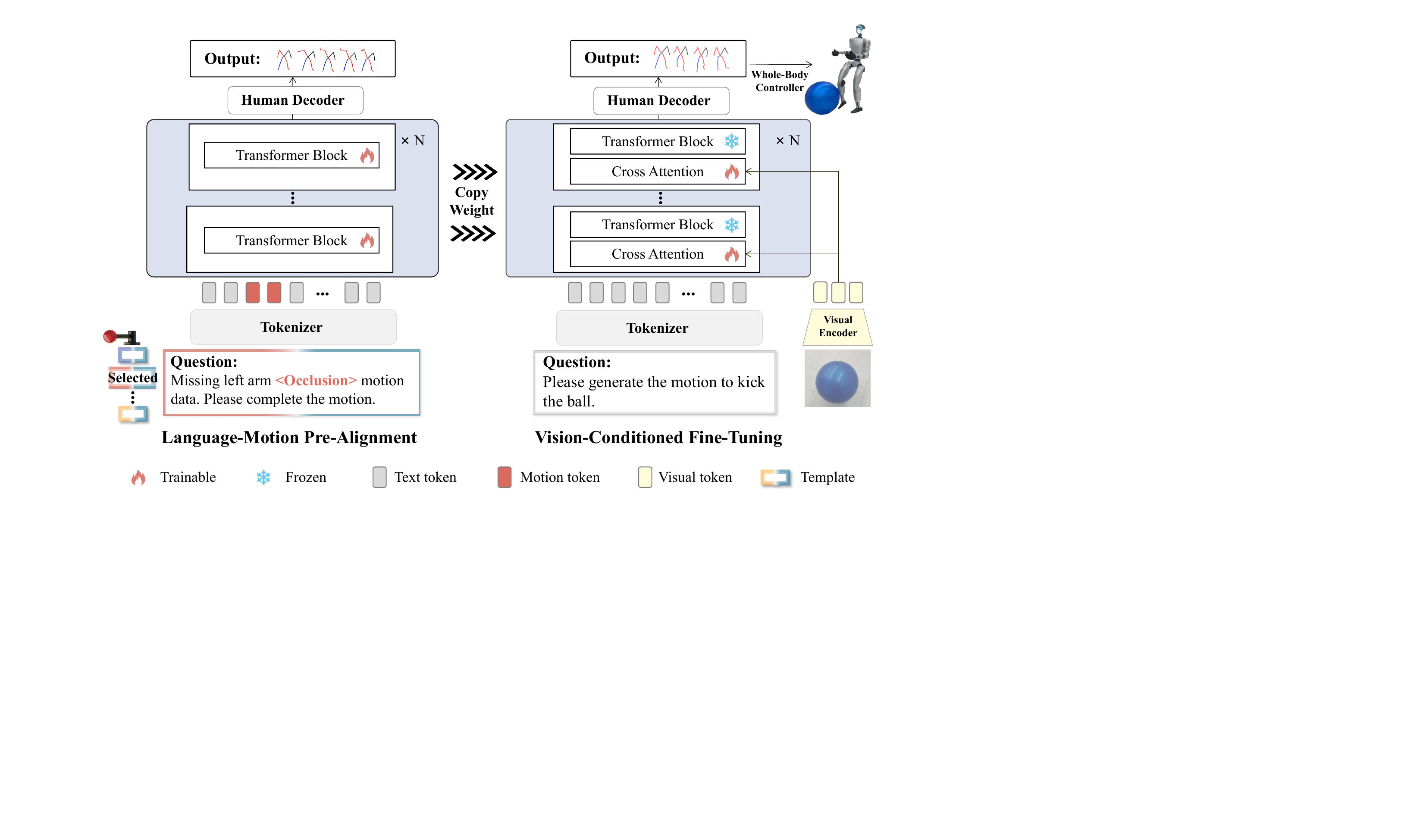}
   \caption{\textbf{Overview of Humanoid-VLA}. Humanoid-VLA includes two main parts: language-Motion Pre-alignment and vision-conditioned fine-tuning. }
   \label{fig:model}
   \vspace{-1.5em}
\end{figure*}

\textbf{Our solution.}
We present Humanoid-VLA, the first VLA model for humanoid robots, seamlessly integrating language understanding, scene perception, and motion control into a unified system to address previous limitations in humanoid control. Next, we demonstrate the framework from two main parts: \textbf{Language-Motion Pre-Alignment} and \textbf{Vision-conditioned Fine-tuning}.

\subsection{Language-Motion Pre-Alignment}
In this stage, we align non-egocentric human motion data with language descriptions. This alignment enables the model to learn motion patterns and action semantics from non-egocentric data sources, laying a robust foundation for motion generation without requiring egocentric visual input.

\begin{figure*}
  \centering
   \includegraphics[width=1\linewidth]{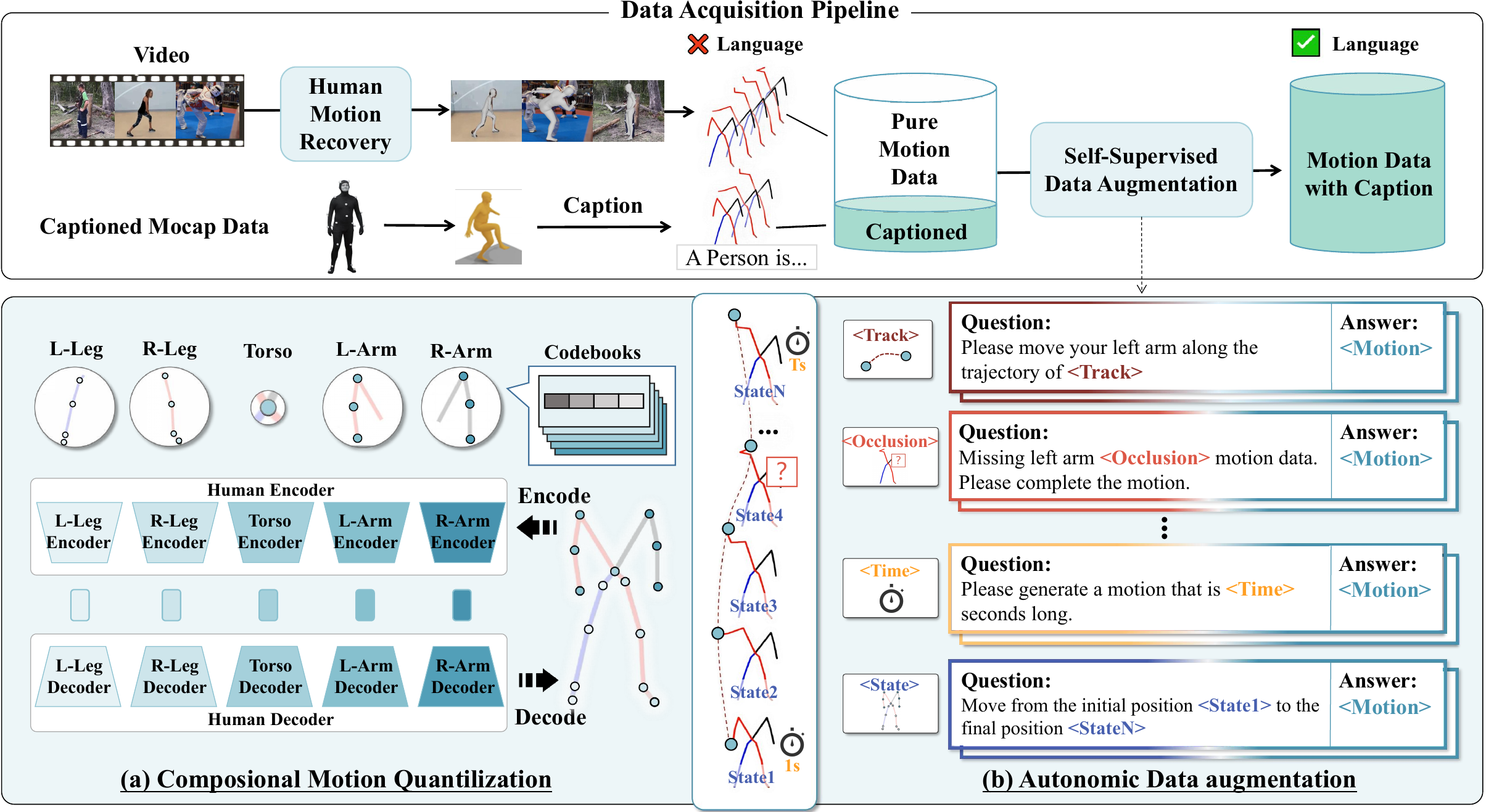}
   \caption{\textbf{Data Acquisition Pipeline}. We propose a cost-effective, self-supervised data augmentation approach that converts abundant pure motion data from videos into annotated motion data with captions. The framework consists of two key modules: a compositional motion quantization method and an autonomous data augmentation approach, which together enable scalable expansion of the dataset.}
   \label{fig:templtes}
   \vspace{-1.5em}
\end{figure*}

\subsubsection{Data Acquisition}
\begin{table}
\centering
\resizebox{1\linewidth}{!}{
\begin{tabular}{@{}c|cc|ccc@{}}
\toprule
Category  & Text & Motion & Clips & Frames & Hours \\
\midrule
Motion capture  &\checkmark&\checkmark & 29K & 0.3M & 4.1 \\
Online Video &\ding{55}& \checkmark& 0.8M & 541M & 7515.7 \\
{Synthetic Data}  &\checkmark& \checkmark& 100K & 16M & 227.7 \\
% Manual Annotated  & \checkmark& \checkmark& 4K & 0.7M & 10.3 \\
% Manual Annotated & Video &  &  & 4K & 0.7M & 10.3 \\
\midrule
\multicolumn{3}{c}{Total}    & \textbf{0.929M} & \textbf{557.3M} & \textbf{7790.2} \\
% Total (A)&  &  &  & \textbf{20.7M} & \textbf{560M} & \textbf{7790.2}\\
\bottomrule
\end{tabular}
}
\caption{\textbf{Datasets Statistics}}
\label{tab:data statistics}
\end{table}
\textbf{Limitations of data acquisition.}
Previous studies have predominantly utilized well-curated datasets that pair motion trajectories with language descriptions to train text-conditioned motion generation models. While these datasets facilitate effective training, they are limited in both quantity and diversity, which constrains their ability to achieve better alignment. In contrast, large-scale online video datasets (as shown in Table~\ref{tab:data statistics}) offer abundant and diverse motion data. However, the absence of corresponding language annotations significantly limits their applicability for this task.

Recent efforts to address this bottleneck have focused on annotating large-scale video datasets manually or using video large language models (VLLMs)\cite{zhang2023video}. However, manual labeling is prohibitively expensive, and VLLMs often produce noisy, incomplete, or imprecise annotations due to their inability to capture fine-grained motion details or describe complex actions. These limitations undermine the effectiveness of the resulting datasets for aligning language and motion.

\textbf{Self-supervised data augmentation.}
Instead of relying on explicit motion descriptions, we propose a cost-effective annotation method by designing various self-supervised tasks directly derived from motion data. For instance, one representative approach involves temporarily masking specific body joints within motion sequences and training the model to reconstruct the occluded movements. Instructional prompts such as "missing left arm <Occlusion> motion data. Please complete the motion" can be generated for these tasks, paired with the corresponding ground truth motions as target outputs. This automatic approach eliminates the need for explicit annotations and is more accurate than adding extra annotation for motion data from video sources.

Next, we explain how this is achieved through two key modules:
\textbf{compositional motion quantitation} and  \textbf{automatic data augmentation.}

\textbf{Compositional motion quantitation.}
As shown in Figure~\ref{fig:model}, we propose a decompositional compression method for body pose representation.
Specifically, we decompose each body pose into five body-based tokens corresponding to five distinct parts: the left leg, right leg, torso, left arm, and right arm. 
We independently train each encoder $\mathcal{E}_b$ and its corresponding codebook $V_b$ for each body part to compress the body part data at time $t$, denoted as $c_t$, into a quantized representation $z_t \in \mathbb{R}^{5}$. 

Formally, we define the motion encoder as $\mathcal{E}_m=\{\mathcal{E}_b\}_{b=1}^5$, which compresses $c_t$ into $z_t$.
\begin{equation}
    \hat{z_t} = \mathcal{E}_m(c_t),
\end{equation}
where $\hat{z_t}=\{\hat{z_b}\}_{b=1}^5$ is the collective discrete vector obtained from $\mathcal{E}_m$, which are the most similar elements to the quantization of $c_t$ in vocabulary $V_m=\{V_b\}_{b=1}^5$. Similar to the encoder, we employ a motion decoder to project the latent variable back onto the action space:
\begin{equation}
    \hat{c_t} = \mathcal{D}_m(\hat{z_t}).
\end{equation}

The optimize goal $\mathcal{L}_{hvq}$ can be expressed as the combination of reconstruction loss $\mathcal{L}_{\text{rec}}$, embedding loss $\mathcal{L}_{\text{emb}}$ and commitment loss $\mathcal{L}_{\text{com}}$:
\begin{equation}
      \mathcal{L}_{hvq} = \underbrace{\|{c_t} - \hat{{c_t}}\|_2}_{\mathcal{L}_{\text{rec}}} + \underbrace{\|\text{sg}({z_t})-\hat{{z_t}}\|_2}_{\mathcal{L}_{\text{emb}}} + \underbrace{\|{z_t}-\text{sg}(\hat{z_t})\|_2}_{\mathcal{L}_{\text{com}}}.
\end{equation}

This compositional encoding method is crucial, allowing for flexible editing of motion sequences. The advantage of decomposing a body pose into multiple parts and encoding them separately lies in that we can form flexible operations on the motion sequence at the token level. For instance, we can replace, perturb, or rearrange the tokens corresponding to specific body parts to generate new motion patterns. This flexibility significantly enhances control over motion data, laying the foundation for further task design.

\textbf{Automatic data augmentation. }
As illustrated in Figure~\ref{fig:templtes}, we introduce four types of augmentations—<Track>, <Time>, <Occlusion>, and <State>—to extract diverse features from raw motion data. For example, in the <Track> augmentation, we isolate the temporal trajectory of a specific joint (e.g., the root joint) and encode it as a corresponding motion token. To create meaningful question-answer pairs, we pair this motion feature with an instructional prompt, such as “Please move your center position along the trajectory of <Track>,” while using the complete motion sequence as the answer. This approach effectively augments datasets that initially lacked linguistic annotations, enabling their use in tasks requiring text-motion alignment.

\textbf{Discussion. }
This method presents several key advantages. 
1) It is highly flexible and extensible: augmentation types like <Track> can be combined with other conditions (e.g., <Time>) to create more complex tasks, while linguistic diversity can be further enriched by rephrasing the same instruction through tools like GPT-4\cite{achiam2023gpt}.
2) The framework leverages motion data's inherent temporal and spatial dynamics, allowing models to learn richer and more robust motion-language relationships.
3) Lastly, the use of interleaved datasets enhances cross-modal alignment by incorporating both motion and text in inputs and outputs. As demonstrated by prior work such as VILA~\cite{lin2024vila}, such training paradigms enable models to better capture the interplay between motion and language without compromising performance on their original tasks.

Using this augmentation approach, we collect the largest motion-language interleaved dataset to date, with a scale that is 25 times larger than previous work~\cite{mao2024learning}. This effectively addresses the data scarcity issue for training foundational human motion models.

\subsubsection{Training.}
When we acquire enough data with language annotations, we still need to consider the quality of raw motion data from video sources.
Therefore, we divide our whole training process into two stages. First, we leverage low-quality data to establish initial alignment between motion and language. Even if they are not precise, the large-scale data could also lay a foundation. 
Later, we continue to train the model using a smaller but high-quality dataset from Mocap, ensuring that it conforms to proper human kinematics.

\textbf{Details.}
We utilize LLMs to map input conditions to generate motion sequences effectively. Our data augmentation approach and compositional motion encoding allow LLMs to seamlessly embed motion conditions into input descriptions.
For instance, an instruction \( l_t \) for motion generation can be structured as: "Plan a sequence of actions ending with <State> over <Time> seconds." Here, <State> corresponds to the discrete action representation token \( z_t \), which is derived from the motion pose \( c_t \) at timestep \( t \) in the motion sequence, while <Time> specifies the motion duration. By unifying the motion codebook \( V_m \) and the language codebook \( V_l \) into a shared vocabulary \( V = \{V_l, V_m\} \), we can encode the instruction \( l_t \) alongside the motion representations \( z_t \) and temporal representations \( d_t \) as language tokens \( X_d = \{x_d^i\}_{i=1}^N \), where \( x_d \in V \) and \( N \) represents the length of the input description. This transformation makes the combined motion and temporal data compatible with LLMs, enabling precise and flexible input encoding.

\textbf{Loss function.}
Motion generation can thus be framed as an autoregressive process that predicts the dictionary index of the next action token, ultimately producing the final motion output \( X_o = \{x_o^i\}_{i=1}^L \), where \( x_o \in V \) and \( L \) denotes the output sequence length. The training objective is defined as maximizing the log-likelihood of the data distribution:
\begin{equation} 
\mathcal{L}_\text{LLM} = -\sum_{i} \log p(x_o^i \mid x_o^{<i}, x_d).
\end{equation}
Finally, the predicted discrete motion sequence \( \hat{z_t} \) can be derived from the LLM's output sequence \( X_o \) through vocabulary mapping. This sequence can then be used to reconstruct the final predicted motion \( S = \{s_t\}_{t=1}^{T} \), where \( T \) represents the length of the motion sequence.

\subsection{Vision-Conditioned Fine-Tuning}
Visual information provides humanoids with detailed object-aware insights, helping them not only understand how to act but also decide what actions to take. While previous research has trained humanoids using large datasets of human motion, the lack of egocentric visual data limits their ability to react based on autonomous perception. To address this, we collect real-world motion capture data paired with egocentric visuals, enabling the transfer of learned motion knowledge to real-world, visually grounded scenarios.

\textbf{Details.}
We copy and freeze the transformer layers from the language-motion pre-alignment phase to integrate visual information with language descriptions. Additionally, we introduce a vision encoder and utilize cross-attention layers to fuse visual features \( X_v \) with language features \( X_d \) into a unified embedding \( X_u \). 
Specifically, the decoder comprises \( L \) layers, with the \( l \)-th layer consisting of a copied transformer decoder layer and a cross-attention layer. In the cross-attention layer, the tokenized language tokens \( X_d^l \) are used as the query, while the encoded visual tokens \( X_v^l \) serve as both the key and the value:
\begin{equation}
\label{fig:overall}
\begin{aligned}
    Q_l = X_d^l W_{Q}^l,\ \
    K_l = X_v^l W_{K}^l,\ \
    V_l = X_v^l W_{V}^l, 
\end{aligned}
\end{equation}
\begin{equation}
\begin{aligned}
{X_u^l} = \text{Softmax}(\frac{Q_lK_l^T}{\sqrt{D}})V_l,
\end{aligned}
\end{equation}
where $D$ represents the hidden dimension size, $W_{Q}^l\in\mathbb{R}^{D_d\times D}$ represents the linear transformation matrix of language token, and $W_{K}^l, W_{V}^l\in\mathbb{R}^{D_v\times D}$ represents the transformation of visual tokens.

\textbf{Loss function.} Here, we optimize the model in the same way as the former language-action pre-alignment phase.

\subsection{Whole-Body Controller}

\textbf{Details.}
Once the two training phases are completed, the model can be integrated with a whole-body controller to enable control of a humanoid robot. 
The whole-body controller $\mathcal{P}$ is essentially a goal-conditioned RL policy that maps human motion onto the joints of a humanoid robot $j_t\in \mathbb{R}^{24}$. We define a reward strategy $\mathcal{R}$, which takes the observation $\mathcal{O}$ and the given goal $\mathcal{G}$ as input, and outputs the target positions for the proportional-derivative (PD) controller in the action space $\mathcal{A}$. We use the proximal policy optimization (PPO) \citep{schulman2017proximal} to maximize the accumulated reward.

%% file: includes/5_exps.tex
\section{Experiments}
We evaluate the proposed \textbf{Humanoid-VLA} in terms of its ability to enable universal humanoid control.
We structure the experiments to answer
the following questions:
\textbf{1) RQ1}: Does \textbf{Humanoid-VLA} generate kinematically accurate and physically plausible motions?
\textbf{2) RQ2}: How effective is the humanoid control with vision integration?

\subsection{Evaluation on motion generation}
In this section, we take a comprehensive evaluation of the quality of the pose trajectory generated by the model.
To comprehensively demonstrate the effectiveness of our approach, we access motion quality from two perspectives:
\textbf{1) Kinematic fidelity:}
This metric evaluates the kinematic performance, measuring positional changes without considering physical dynamics.
Following~\cite{mao2024learning}, we evaluate our model on the standard text-to-motion (T2M) task, which generates motion sequences based on textual action descriptions. It highlights the model's core capability of translating natural language into human motion.  

\textbf{2) Physical plausibility:}
Unlike the above metric, this evaluation assesses the physical feasibility of generated poses in real-world environments.
Beyond standard T2M tasks, we evaluate our model on more challenging scenarios that exceed the capabilities of existing models, particularly tasks incorporating diverse input conditions such as joint trajectories. This comprehensive assessment demonstrates the robustness and versatility of the model across a broad spectrum of applications.

\vspace{-4pt}
\subsubsection{Kinematic Fidelity }
\vspace{-4pt}
\textbf{Setup.} 
We evaluate the motion quality using a widely used dataset HumanML3D~\cite{guo2022humanml3d} and our collected dataset Humanoid-S, which contains manually annotated action descriptions for human pose extracted from 4646 video clips.
While HumanML3D focuses on basic locomotion patterns such as running, swimming, and dancing, Humanoid-S encompasses more complex human actions.
We choose the whole testing dataset and randomly select one textual description per clip to serve as the input for evaluation. 
For a fair comparison, we evaluate all models using 15 joints consistent with our model's configuration, selected for their presence in both humans and humanoid robots to enhance generalizability.

\begin{table}
\centering
\resizebox{0.9\linewidth}{!}{%
\begin{tabular}{@{}lccccccc@{}}
\toprule
\multirow{2}{*}{Types} &  
\multirow{2}{*}{Input} & 
\multicolumn{4}{c}{Accuracy}
\\ \cmidrule(lr){3-6} 
&
&$E_{\text{mpjpe}}^{g}\downarrow$ 
% &$E_{\text{mpjpe}}^{l}\downarrow$ 
&$E_{\text{mpjpe}}^\text{pa}\downarrow$ 
& $E_{\text{accel}} \downarrow$ 
& $E_{\text{vel}}\downarrow$ 
\\ \midrule
\multirow{4}{*}{Easy}&

{D} &
36.13 & 1.53 & 34.42 & 18.73
\\
&
{T} &
 36.57 & 1.48 & 35.10 & 18.53
\\
&
{A} &
 39.02  & 1.32 & 34.32 & 17.91
\\
& 
 {$S_n$} &
 36.29 & 1.55 & 34.93 & 18.88
\\
\midrule

\multirow{3}{*}{Medium}&
{D + T} &
 {31.07} & {1.18}  & {27.84} & {14.76}
\\
&
 {D + A} &
 36.98& 1.30& 34.87& 18.16
\\
&
 {D + $S_n$} &
 35.75 & {1.18} &33.41 &17.18
\\
\midrule
\multirow{1}{*}{Hard}& 
 {D + $S_1$ + $S_N$} 
 & 37.14 & 1.34 & 34.69 & 18.08
\\
   \bottomrule
\end{tabular}
}
\caption{\textbf{Physical plausibility of generated motion under versatile conditions.} Humanoid-VLA provides four conditional input types: motion description (\textbf{D}), motion time duration (\textbf{T}), motion sequence with absent body parts (\textbf{A}), and motion state (\textbf{$S_n$}) at time n within total N timesteps.
Based on input combinations, we establish three tiers of motion generation tasks with increasing complexity.}
\label{tab:controllable}
\vspace{-1.em}
\end{table}

\begin{table}
\centering
\resizebox{1\linewidth}{!}{%
\begin{tabular}{@{}ccccccc@{}}
\toprule
\multirow{2}{*}{Method} & 
\multicolumn{2}{c}{HumanML3D}& \multicolumn{2}{c}{Humanoid-S}\\
\cmidrule(lr){2-3}
\cmidrule(lr){4-5}  
% & MM Dist$\downarrow$ 
& FID$\downarrow$ 
% & MModality$\uparrow$
& DIV$\uparrow$ 
% & MMDist$\downarrow$ 
& FID$\downarrow$ 
% & MModality$\uparrow$
& DIV$\uparrow$ 
\\ \midrule
MDM&
% $\boldsymbol{3.740}^{\pm.095}$ &
${0.889}^{\pm.026}$ &
${3.855}^{\pm.053}$ &
${2.351}^{\pm.590}$ &
${4.111}^{\pm.261}$ &
\\
T2M-GPT&
% $4.512^{\pm.165}$ &
${0.531}^{\pm.020}$ &
${4.555}^{\pm.058}$ &
${1.101}^{\pm.189}$ &
${4.199}^{\pm.218}$ &
\\\midrule
% 10 epoch微调版本结果
\modelname &
% $4.270^{\pm.206}$ &
$\boldsymbol{0.467}^{\pm.018}$ & 
$\boldsymbol{4.585}^{\pm.086}$ &
$\boldsymbol{1.037}^{\pm.147}$ & 
$\boldsymbol{4.466}^{\pm.213}$ &
\\
 \midrule
\end{tabular}%
}
\caption{
\textbf{Kinematic fidelity of generated motion in HumanML3D and Humanoid-S.} We use FID score and Diversity to evaluate the quality of the motion generated by the model, where bold values indicate the best results.
}
\label{tab:t2m}
\vspace{-1.5em}
\end{table}

\textbf{Metrics.}
We follow the evaluation framework from ~\cite{guo2022fid}, utilizing two established metrics to evaluate the quality of motion generation: \textbf{(1) FID} measuring distribution similarity between generated and real motions, and \textbf{(2) Diversity} quantifying action variation degree, and calculating the average Euclidean distance between 200 randomly generated motions.
Lower FID indicates better distribution matching and higher DIV scores reflect superior diversity.

\textbf{Baselines.}
We consider two baselines commonly used in humanoid control:
{\textbf{(1) MDM}~\cite{tevet2023human}}: A diffusion-based generation model that utilizes a classifier-free paradigm to produce natural and diverse motions.
{\textbf{(2) T2M-GPT}~\cite{zhang2023t2m}}: A transformer-based generation model that combines VQ-VAE\cite{van2017neural} with an autoregressive approach to generate human motions from text.

\textbf{Implementation details.}
We utilize Llama3-70B~\cite{dubey2024llama} as the foundation model. In the training phase, warm up ratio is set at 0.01, with learning rate configured at 2e-5, and the cosine learning scheduler. The batch size per device is set to 4. For the encoder of each body part, its codebook size is set to 1024. We conduct model training using 8 NVIDIA H100 GPUs through {216} hours.

\begin{figure*}[t]
  \centering
   \includegraphics[width=1\linewidth]{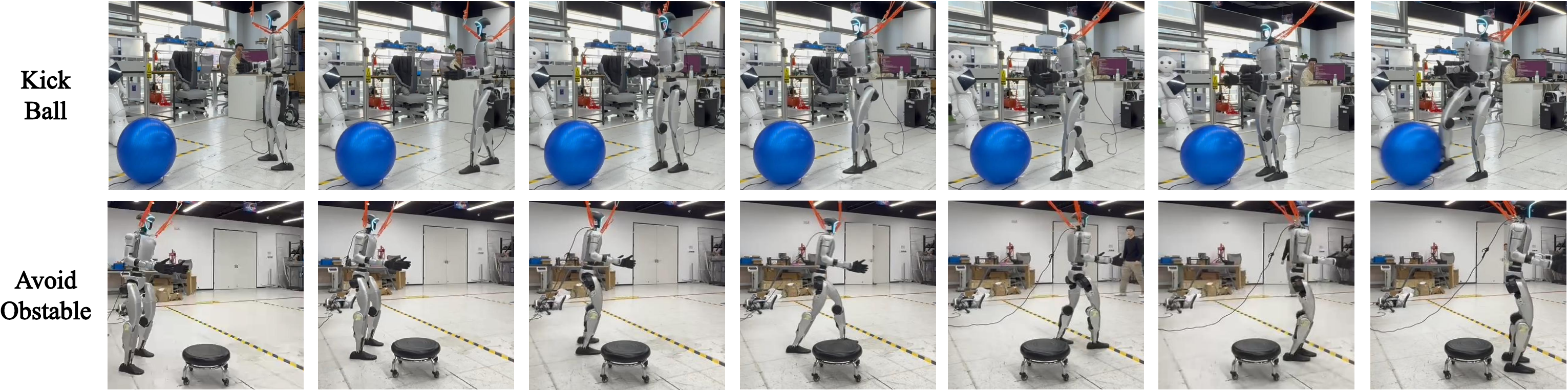}
   \caption{\textbf{Robot experiments in real world}. \modelname demonstrates its ability to interact with objects, showcasing robust performance in real-world environments. The humanoid model successfully executes precise object-kicking tasks and avoids obstacle task in real-world scenarios.}
   \label{fig:real}
   \vspace{-1.5em}
\end{figure*}

\textbf{Results.}
The comparative evaluation results between \modelname and the baseline models are presented in Table~\ref{tab:t2m}. On the HumanML3D dataset, \modelname achieves a significant FID score of 0.467, representing substantial improvements of 47.5\% and 12\% over MDM and T2M-GPT respectively, which indicates its superior capability in capturing real motion distributions. Furthermore, \modelname demonstrates remarkable performance in motion diversity, attaining a diversity score of 4.466 on the Humanoid-S dataset, outperforming MDM by 6\%. This achievement is particularly noteworthy as it reflects the model's ability to generate diverse motions under challenging linguistic constraints. The comprehensive experimental results demonstrate that \modelname excels in high-quality action generation, establishing its effectiveness in text-to-motion synthesis tasks.

\begin{table}
\centering
\resizebox{1\linewidth}{!}{%
\begin{tabular}{@{}cccccccc@{}}
\toprule
& 
Low-quality data &
\multicolumn{1}{c}{High-quality Data} &
\multirow{2}{*}{FID$\downarrow$} &
\multirow{2}{*}{DIV$\uparrow$ }
% \multicolumn{3}{c}{Motion-to-Text}& 
\\ \cmidrule(lr){2-2}\cmidrule(lr){3-3}& w aug & w aug   
% MModality$\uparrow$

% & MMDist$\downarrow$ 
\\ \midrule
&\checkmark & &
${0.698}^{\pm.037}$ & 
${4.576}^{\pm.098}$ &\\
& &\checkmark &
${0.557}^{\pm.016}$ & 
${3.867}^{\pm.062}$ &\\
 & \checkmark&\checkmark &
$\boldsymbol{0.467}^{\pm.018}$ & 
$\boldsymbol{4.585}^{\pm.086}$ &\\
   \bottomrule
\end{tabular}%
}
\caption{\textbf{Ablation on data augmentation.} Here, low-quality data refers to motion data extracted through human motion recovery, which tends to lack precision. In contrast, high-quality data refers to motion data obtained directly from physical devices, ensuring greater accuracy.}
\label{tab:aug}
\vspace{-1.5em}
\end{table}

\subsubsection{Physical Plausibility }
\textbf{Setup.} 
We evaluate this metric in the IsaacGym physics simulator~\cite{makoviychuk2021isaac}.
Following~\cite{he2024omnih2o,ji2024exbody2}, we assess the humanoid's tracking accuracy in executing the model-generated kinematic trajectories to quantify physical plausibility.

\textbf{Metrics.}
Our metrics are designed across two dimensions: (1) \textit{\textbf{State-related.}} The global Mean Per-Joint Position Error (MPJPE) $E_{\text{mpjpe}}^{g}$ (mm) quantifies the average positional error of individual joints. The Procrustes-aligned MPJPE (PA-MPJPE) $E_{\text{mpjpe}}^\text{pa}$ (mm) eliminates global scale and rotational discrepancies to assess shape accuracy specifically. (2) \textit{\textbf{Transition-related.}} We evaluate acceleration error $E_{\text{accel}}$ (mm/s²) and velocity error $E_{\text{vel}}$ (mm/s) metrics to assess physical plausibility by computing the average joint-level distances in acceleration and velocity respectively.
For all metrics, lower values correspond to better performance.

\textbf{Baselines.}
As these conditional motion tasks are uniquely solvable by our model, conventional baselines are not applicable. Therefore, we focus on evaluating our approach independently, adopting the tracking error widely accepted in ~\cite{ji2024exbody2} to evaluate the effectiveness of our method.

\textbf{Results.}
As shown in Table \ref{tab:controllable}, our RL policy achieves robust motion imitation joint control with mean position errors \(E_{\text{mpjpe}}^g\) consistently below 40 mm, and minimum score 31.07mm under medium difficulty with caption and time conditions. The policy achieves remarkably low errors in pose accuracy $E_{\text{mpjpe}}^\text{pa}$ at {1.18mm}, acceleration $E_{\text{accel}}$ at {27.84mm}, and velocity $E_{\text{vel}}$ at {14.76mm}, demonstrating smooth and physically consistent motion generation. This experiment validates our method's ability to generate high-quality motions across diverse control conditions while preserving physical plausibility and control fidelity.

\textbf{Ablation on data augmentation.} 

As shown in Table \ref{tab:aug}, incorporating extensive video motion data reduces the FID from 0.557 to 0.467, representing a 16\% improvement. This significant enhancement demonstrates that large-scale motion data extracted from videos strengthens the alignment between motion and language. 
We can still achieve comparable results Even with low-quality mocap data for fine-tuning. These two points strongly validate the significance of incorporating video data to expand the training process.
These findings underscore the effectiveness of our self-supervised data augmentation strategy.

\vspace{-10pt}
\subsection{Evaluation on vision integration }

\vspace{-5pt}
\textbf{Experimental setup.}

We evaluate the performance of our \modelname model in real-world environments utilizing visual information. An RGB camera is employed to capture first-person view images, and experiments are conducted using the Unitree G1 robot across four task categories: upper-body motion, lower-body motion, full-body motion, and object interaction. These tasks, such as approaching targets, kicking a ball, and obstacle navigation, require visual guidance for accurate positioning, aiming to validate the effectiveness of our VLA approach.

\textbf{Results.}
For each task category, we carefully select 4 representative tasks and evaluate 10 tests on each task, using success rate as the evaluation metric.
Our humanoid-VLA model shows great performance in various object interaction tasks shown in Table ~\ref{tab:vis_SR}.
Selected tasks are shown in Figure~\ref{fig:real}. 
In the "kick ball" task, our humanoid model enables the robot to effectively utilize visual information to accurately approach the object and execute a kicking motion. 
In the "avoid obstacles" task, our robot successfully navigates around obstacles to reach the desired target position. 
These results demonstrate that our VLA model effectively leverages visual information to generate appropriate motions.

\begin{table}[ht]
  \centering
  \resizebox{0.45\linewidth}{!}{
  \begin{tabular}{@{}c|cc@{}}
    \toprule
    \textbf{Task} & \textbf{\makecell[c]{SR}} \\
    \midrule
 Turn to an object & 10/10 \\
 Hold an object  & 9/10 \\

 Wave to people & 10/10 \\

Avoid an obstacle  & 9/10 \\  
Jump over an object & 9/10 \\  
Dance with a partner  & 8/10 \\ 
 Punch an obstacle & 10/10 \\ 
 Kick a ball  & 9/10 \\

    \bottomrule
  \end{tabular}
  }
  \caption{\textbf{Evaluation on Vision Integration.} }
  \label{tab:vis_SR}
  \vspace{-2em}
\end{table}

%% file: includes/6_conclus.tex
\section{Conclusion}
This paper introduces Humanoid-VLA, a novel framework designed to address the challenges of humanoid robot control with egocentric visual integration. The framework aligns language and motion using human motion datasets, enables context-aware motion generation through cross-attention mechanisms, and tackles data scarcity using self-supervised pseudo-annotations. Built on whole-body control architectures, Humanoid-VLA facilitates adaptive object interaction and exploration with enhanced contextual understanding.
The effectiveness of Humanoid-VLA has been validated through evaluations of motion generation quality and execution success rates on real humanoid robots, demonstrating high executability. In the future, we aim to enhance the success rate of humanoid robots in performing more complex loco-manipulation tasks.

%% file: includes/8_appendix.tex
\newpage
\appendix
\onecolumn

\section{Data collection}

Our motion dataset is derived from three primary sources: Motion Capture, Online Videos, and Synthetic Data. Specifically, the motion capture data is sourced from the open-source AMASS dataset, which includes textual annotations. Synthetic Data is generated by inputting random text into an open-source motion model to produce corresponding movements. Online Videos are collected from the web, with human motions extracted using the method described in ~\cite{wang2025tram}.

It is important to note that we select 15 universal humanoid joint points from the standard 22 SMPL joints in these open-source datasets. The framework enhances flexibility and extensibility by seamlessly integrating augmentation types with other conditions to create complex tasks, while advanced tools such as GPT-4 expand linguistic diversity through instruction rephrasing. By leveraging the inherent temporal and spatial dynamics of motion data, it enables models to learn more comprehensive and robust motion-language relationships. Additionally, improved cross-modal alignment is achieved through interleaved datasets that incorporate both motion and text, allowing models to better capture the interplay between motion and language.

\section{Data templates}

\begin{table}[ht]
  \centering
  \resizebox{1\linewidth}{!}
  {
  \begin{tabular}{@{}c|l@{}}
    \toprule
    \textbf{Task} & \textbf{\makecell[c]{Template descriptions}} \\
    \midrule
\multirow{9}{*}{\texttt{<Occlusion>} $\to$ \texttt{<Motion>}} 
& \textbf{Your help is crucial for us. Please complete the missing left leg data: \texttt{<Occlusion>}.}\\
&\quad1. \texttt{<Occlusion>} is part of the left leg motion. Please predict its complete data.\\
&\quad2. The left leg data: \texttt{<Occlusion>} is incomplete. Please help us fill in this part.\\
&\quad3. Please provide the complete left leg data: \texttt{<Occlusion>}.\\
&\quad...\\
&\quad N. To perfect the motion, we need the complete left leg data: \texttt{<Occlusion>}. Please assist us.\\
\\
& Your help is crucial for us. Please complete the missing right leg data: \texttt{<Occlusion>}. (\texttimes N)\\
& \texttt{<Occlusion>} is part of the left arm motion. Please predict its complete data. (\texttimes N)\\
& The right arm data: \texttt{<Occlusion>} is missing. Could you help us fill in this part? (\texttimes N)\\
\midrule
\multirow{3}{*}{\texttt{<Track>} $\to$ \texttt{<Motion>}} & Please ensure your root moves along \texttt{<Track>} for the next action. (\texttimes N)\\
& Please move your left hand according to the trajectory of \texttt{<Track>}. (\texttimes N) \\
& Please keep the movement of your right hand consistent with \texttt{<Track>}. (\texttimes N)\\
\midrule
\multirow{3}{*}{\texttt{<Motion>} $\to$ \texttt{<Track>}} 
& I would like a detailed analysis of the trajectory of the central position in this action: \texttt{<Motion>}. (\texttimes N)\\
& Tell me the path of the left hand: \texttt{<Motion>}. (\texttimes N)\\
& I would like a detailed analysis of the trajectory of the right hand in this action: \texttt{<Motion>}. (\texttimes N)\\

\midrule
\multirow{2}{*}{\texttt{<Time>} $\to$ \texttt{<Motion>}} 
& Can you make a motion that lasts for \texttt{<Time>} frames, with a certain percentage of variability? (\texttimes N)\\
& Show me a motion that is longer than \texttt{<Time>} seconds in duration. (\texttimes N)\\

\midrule
\multirow{2}{*}{\texttt{<Motion>} $\to$ \texttt{<Time>}} 
& Calculate the frame duration for \texttt{<Motion>}'s poses. (\texttimes N)\\
& Compute the duration in seconds for \texttt{<Motion>}'s poses. (\texttimes N)\\

\midrule
\multirow{3}{*}{\texttt{<State>} $\to$ \texttt{<Motion>}} 
& Randomly generate an entire action sequence from \texttt{<State1>}. (\texttimes N)\\
& Create actions randomly using the last state \texttt{<StateN>}. (\texttimes N)\\
& Move from the initial position \texttt{<State1>} to the final position \texttt{<StateN>}. (\texttimes N)\\
\midrule

\multirow{2}{*}{\texttt{<Caption>}+\texttt{<Occlusion>} $\to$ \texttt{<Motion>}} 
 & Explain the state before executing \texttt{<Motion>} actions. (\texttimes N)\\
& Explain the final conditions following \texttt{<Motion>} actions. (\texttimes N)\\
\midrule

\multirow{2}{*}{\texttt{<State>}+\texttt{<Time>} $\to$ \texttt{<Motion>}} &
Design a full action sequence culminating in \texttt{<StateN>} across \texttt{<Time>} frames. (\texttimes N)\\
& Plan a sequence of actions ending with \texttt{<StateN>} over \texttt{<Time>} seconds. (\texttimes N)\\
\midrule

\multirow{4}{*}{\texttt{<Occlusion>}+\texttt{<Caption>} $\to$ \texttt{<Motion>}} &
 To complete the \texttt{<Caption>} action, we need the missing center motion data: \texttt{<Occlusion>}. Please assist us. (\texttimes N)\\
& To complete the \texttt{<Caption>} action, we need the missing left leg motion data: \texttt{<Occlusion>}. Please assist us. (\texttimes N)\\
& To finish the \texttt{<Caption>} action, we need to fill in the missing left arm motion data: \texttt{<Occlusion>}. Please assist us. (\texttimes N)\\
& Your assistance is crucial. Please help us complete the missing right arm data: \texttt{<Occlusion>} for the \texttt{<Caption>} action. (\texttimes N)\\
\midrule

\multirow{3}{*}{\texttt{<Track>}+\texttt{<Caption>} $\to$ \texttt{<Motion>}} &
 Please keep your root on the trajectory of \texttt{<Track>} while performing the actions described by \texttt{<Caption>}. (\texttimes N)\\
& While your left hand is moving along \texttt{<Track>}, perform the action described by \texttt{<Caption>}. (\texttimes N)\\
& Generate an action according to the description \texttt{<Caption>} while ensuring your right hand remains on the trajectory of \texttt{<Track>}. (\texttimes N)\\
\midrule

\multirow{3}{*}{\texttt{<State>}+\texttt{<Track>}+\texttt{<Caption>} $\to$ \texttt{<Motion>}} &
Starting from \texttt{<State1>}, follow the direction of your root guided by \texttt{<Track>} to perform the dynamic described by \texttt{<Caption>}. (\texttimes N)\\
& Starting from \texttt{<State1>}, your right hand needs to follow the path of \texttt{<Track>} and complete the dynamic described in \texttt{<Caption>}. (\texttimes N)\\
& Starting from \texttt{<State1>}, follow the direction of your root guided by \texttt{<Track>} to perform the dynamic described by \texttt{<Caption>}. (\texttimes N)\\
\midrule
\multirow{1}{*}{\texttt{<Motion>}+\texttt{<Time>}+\texttt{<Caption>} $\to$ \texttt{<Motion>}} &
Produce an action sequence for \texttt{<Caption>} incorporating a partial motion sequence described as \texttt{<Motion>} spanning \texttt{<Time>} frames. (\texttimes N)\\
    \bottomrule
  \end{tabular}
  }
  \caption{\textbf{Examples of conditional language description.} All task descriptions could be expanded $N$ times similar to the first example. }
  \label{tab:app_templates}
\end{table}

The input descriptions for our subtasks are presented in Table ~\ref{tab:app_templates}. Specifically, for each category of subtasks, such as "Complete left leg data," we have developed N variations of expressions. This allows our self-supervised augmentation approach to significantly expand the range of task descriptions, increasing the original 59 subtasks at N times.

\section{Simulation Performance}

\begin{figure*}[h]
  \centering
   \includegraphics[width=1\linewidth]{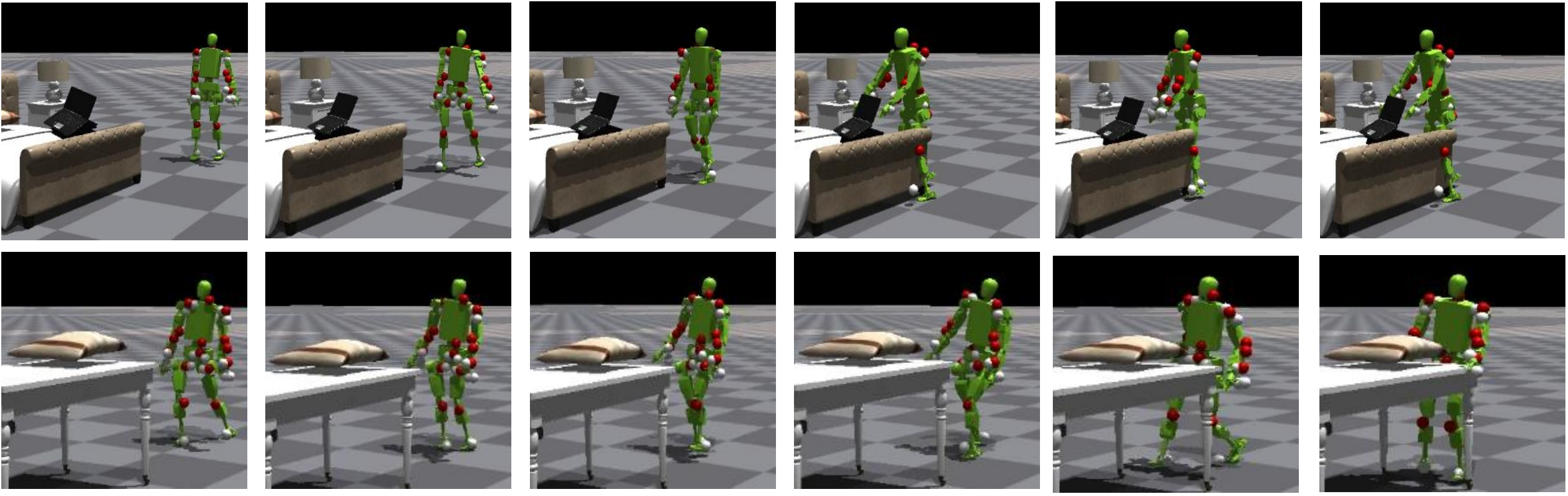}
   \caption{\textbf{Simulation robot experiment}.}
   \label{fig:humanvla}
\end{figure*}

To enhance humanoid robots' environmental interaction capabilities, we selected 2 representative object interaction tasks from the HITR dataset referencing HumanVLA ~\cite{xu2024humanvla}. For each task, we collected comprehensive data including egocentric visual frames, natural language instructions, and robot control signals. Our approach extends beyond real-world scenarios into simulation environments, where we followed HumanVLA's setup to extract 15 key joints from stick figure trajectories for constructing our training dataset. This methodology leverages the abundance of existing stick figure/humanoid datasets, enabling efficient data acquisition through retargeting for both visual and motion aspects. Although joint retargeting to humanoid forms may not match the quality of motion capture data, it presents promising opportunities for continued research using large-scale simulation-based VLA data for downstream task training.

A key strength of our universal framework lies in its demonstrated adaptability across different robot configurations. Unlike previous approaches that are often constrained to specific platforms, our framework successfully generalizes between distinct robot architectures in simulation and real-world implementations. This universal applicability represents a significant advancement over prior work such as HumanVLA, which, while providing excellent policy foundations, was limited in achieving universal humanoid control and real robot deployment. Our framework transcends these limitations by establishing a universal bridge between simulation and physical robot systems, paving the way for truly generalizable humanoid control strategies that can be deployed across diverse platforms and tasks.

\section{More Details}
It is noteworthy that the joint points for motion generation in our model are not the conventional 22 joints, but rather a set of 15 joints that are common to both humanoid robots and humans. This approach ensures universality across different configurations. However, an additional optimization step is required to adapt the poses generated by the large model for downstream applications. Specifically, generating a motion sequence with 15 joints necessitates an optimization process to map it onto the target downstream configuration.

To acquire training data, we extract 15 joint points from standard datasets for training purposes. This ensures that the model learns from a consistent and universally applicable set of joint points.

To enable humanoid robots to execute corresponding motions through joint-point mapping, we employ the Adam optimizer similar to ~\cite{mao2024learning, jiang2024harmon}, which map the positions of the 15 joints from keypoints to the 24 joints of the humanoid robot. By maintaining the end-effector positions as closely aligned as possible with the existing joint node positions, the overall motion pattern of the humanoid robot remains consistent with the keypoint representation.

We define the problem as training a goal-conditioned RL policy $\pi$ that maps human motion onto the joints of a humanoid robot $j_t\in \mathbb{R}^{24}$. We define a reward strategy $\mathcal{R}$, which takes the observation $\mathcal{O}$ and the given goal $\mathcal{G}$ as input, and outputs the target positions for the proportional-derivative (PD) controller in the action space $\mathcal{A}$. We use the proximal policy gradient (PPO) to maximize the accumulated reward.

\section{Limitation}

\textbf{Robustness of the RL Policy}. Although we have developed a general RL policy, its performance lacks sufficient robustness. We plan to further refine the policy to enhance task completion.

\textbf{Limited Availability of High-Quality Data}. The availability of high-quality data is limited, including manually annotated data and execution data from real-world humanoid robots. While we considered leveraging datasets from works like Mimicking-Bench~\cite{liu2024mimicking}, the restricted robot configurations render them unsuitable for general robotics tasks. Consequently, we will undertake the collection of data for general humanoid robot tasks and encourage the community to recognize and address this data scarcity.

\textbf{Training Approach}. Our current training methodology is relatively simple, and not fully leveraging the available data. We have identified several strategies to enhance the training of motion generation models, we will incorporate these techniques into our future work.

%% file: main.bbl
\begin{thebibliography}{43}
\providecommand{\natexlab}[1]{#1}
\providecommand{\url}[1]{\texttt{#1}}
\expandafter\ifx\csname urlstyle\endcsname\relax
  \providecommand{\doi}[1]{doi: #1}\else
  \providecommand{\doi}{doi: \begingroup \urlstyle{rm}\Url}\fi

\bibitem[Achiam et~al.(2023)Achiam, Adler, Agarwal, Ahmad, Akkaya, Aleman, Almeida, Altenschmidt, Altman, Anadkat, et~al.]{achiam2023gpt}
Achiam, J., Adler, S., Agarwal, S., Ahmad, L., Akkaya, I., Aleman, F.~L., Almeida, D., Altenschmidt, J., Altman, S., Anadkat, S., et~al.
\newblock Gpt-4 technical report.
\newblock \emph{arXiv preprint arXiv:2303.08774}, 2023.

\bibitem[Black et~al.(2024)Black, Brown, Driess, Esmail, Equi, Finn, Fusai, Groom, Hausman, Ichter, et~al.]{black2024pi_0}
Black, K., Brown, N., Driess, D., Esmail, A., Equi, M., Finn, C., Fusai, N., Groom, L., Hausman, K., Ichter, B., et~al.
\newblock $pi\_0 $: A vision-language-action flow model for general robot control.
\newblock \emph{arXiv preprint arXiv:2410.24164}, 2024.

\bibitem[Brohan et~al.(2023)Brohan, Brown, Carbajal, Chebotar, Chen, Choromanski, Ding, Driess, Dubey, Finn, et~al.]{brohan2023rt}
Brohan, A., Brown, N., Carbajal, J., Chebotar, Y., Chen, X., Choromanski, K., Ding, T., Driess, D., Dubey, A., Finn, C., et~al.
\newblock Rt-2: Vision-language-action models transfer web knowledge to robotic control.
\newblock \emph{arXiv preprint arXiv:2307.15818}, 2023.

\bibitem[Catalin~Ionescu(2011)]{IonescuSminchisescu11}
Catalin~Ionescu, Fuxin~Li, C.~S.
\newblock Latent structured models for human pose estimation.
\newblock In \emph{International Conference on Computer Vision}, 2011.

\bibitem[Cheang et~al.(2024)Cheang, Chen, Jing, Kong, Li, Li, Liu, Wu, Xu, Yang, et~al.]{cheang2024gr}
Cheang, C.-L., Chen, G., Jing, Y., Kong, T., Li, H., Li, Y., Liu, Y., Wu, H., Xu, J., Yang, Y., et~al.
\newblock Gr-2: A generative video-language-action model with web-scale knowledge for robot manipulation.
\newblock \emph{arXiv preprint arXiv:2410.06158}, 2024.

\bibitem[Cheng et~al.(2024)Cheng, Ji, Chen, Yang, Yang, and Wang]{cheng2024expressive}
Cheng, X., Ji, Y., Chen, J., Yang, R., Yang, G., and Wang, X.
\newblock Expressive whole-body control for humanoid robots.
\newblock \emph{arXiv preprint arXiv:2402.16796}, 2024.

\bibitem[Dai et~al.(2014)Dai, Valenzuela, and Tedrake]{dai2014whole}
Dai, H., Valenzuela, A., and Tedrake, R.
\newblock Whole-body motion planning with centroidal dynamics and full kinematics.
\newblock In \emph{2014 IEEE-RAS International Conference on Humanoid Robots}, pp.\  295--302. IEEE, 2014.

\bibitem[Dantec et~al.(2021)Dantec, Budhiraja, Roig, Lembono, Saurel, Stasse, Fernbach, Tonneau, Vijayakumar, Calinon, et~al.]{dantec2021whole}
Dantec, E., Budhiraja, R., Roig, A., Lembono, T., Saurel, G., Stasse, O., Fernbach, P., Tonneau, S., Vijayakumar, S., Calinon, S., et~al.
\newblock Whole body model predictive control with a memory of motion: Experiments on a torque-controlled talos.
\newblock In \emph{2021 IEEE International Conference on Robotics and Automation (ICRA)}, pp.\  8202--8208. IEEE, 2021.

\bibitem[Ding et~al.(2025)Ding, Zhao, Zhang, Song, Zhang, Huang, Yang, and Wang]{ding2025quar}
Ding, P., Zhao, H., Zhang, W., Song, W., Zhang, M., Huang, S., Yang, N., and Wang, D.
\newblock Quar-vla: Vision-language-action model for quadruped robots.
\newblock In \emph{European Conference on Computer Vision}, pp.\  352--367. Springer, 2025.

\bibitem[Dubey et~al.(2024)Dubey, Jauhri, Pandey, Kadian, Al-Dahle, Letman, Mathur, Schelten, Yang, Fan, et~al.]{dubey2024llama}
Dubey, A., Jauhri, A., Pandey, A., Kadian, A., Al-Dahle, A., Letman, A., Mathur, A., Schelten, A., Yang, A., Fan, A., et~al.
\newblock The llama 3 herd of models.
\newblock \emph{arXiv preprint arXiv:2407.21783}, 2024.

\bibitem[Elobaid et~al.(2023)Elobaid, Romualdi, Nava, Rapetti, Mohamed, and Pucci]{elobaid2023online}
Elobaid, M., Romualdi, G., Nava, G., Rapetti, L., Mohamed, H. A.~O., and Pucci, D.
\newblock Online non-linear centroidal mpc for humanoid robots payload carrying with contact-stable force parametrization.
\newblock In \emph{2023 IEEE International Conference on Robotics and Automation (ICRA)}, pp.\  12233--12239. IEEE, 2023.

\bibitem[Guo et~al.(2020)Guo, Zuo, Wang, Zou, Sun, Deng, Gong, and Cheng]{guo2020action2motion}
Guo, C., Zuo, X., Wang, S., Zou, S., Sun, Q., Deng, A., Gong, M., and Cheng, L.
\newblock Action2motion: Conditioned generation of 3d human motions.
\newblock In \emph{Proceedings of the 28th ACM International Conference on Multimedia}, pp.\  2021--2029, 2020.

\bibitem[Guo et~al.(2022{\natexlab{a}})Guo, Zou, Zuo, Wang, Ji, Li, and Cheng]{Guo_2022_CVPR}
Guo, C., Zou, S., Zuo, X., Wang, S., Ji, W., Li, X., and Cheng, L.
\newblock Generating diverse and natural 3d human motions from text.
\newblock In \emph{Proceedings of the IEEE/CVF Conference on Computer Vision and Pattern Recognition (CVPR)}, pp.\  5152--5161, June 2022{\natexlab{a}}.

\bibitem[Guo et~al.(2022{\natexlab{b}})Guo, Zou, Zuo, Wang, Ji, Li, and Cheng]{guo2022fid}
Guo, C., Zou, S., Zuo, X., Wang, S., Ji, W., Li, X., and Cheng, L.
\newblock Generating diverse and natural 3d human motions from text.
\newblock In \emph{Proceedings of the IEEE/CVF Conference on Computer Vision and Pattern Recognition}, pp.\  5152--5161, 2022{\natexlab{b}}.

\bibitem[Guo et~al.(2022{\natexlab{c}})Guo, Zou, Zuo, Wang, Ji, Li, and Cheng]{guo2022humanml3d}
Guo, C., Zou, S., Zuo, X., Wang, S., Ji, W., Li, X., and Cheng, L.
\newblock Generating diverse and natural 3d human motions from text.
\newblock In \emph{Proceedings of the IEEE/CVF Conference on Computer Vision and Pattern Recognition (CVPR)}, pp.\  5152--5161, June 2022{\natexlab{c}}.

\bibitem[He et~al.(2024{\natexlab{a}})He, Luo, He, Xiao, Zhang, Zhang, Kitani, Liu, and Shi]{he2024omnih2o}
He, T., Luo, Z., He, X., Xiao, W., Zhang, C., Zhang, W., Kitani, K., Liu, C., and Shi, G.
\newblock Omnih2o: Universal and dexterous human-to-humanoid whole-body teleoperation and learning.
\newblock \emph{arXiv preprint arXiv:2406.08858}, 2024{\natexlab{a}}.

\bibitem[He et~al.(2024{\natexlab{b}})He, Luo, Xiao, Zhang, Kitani, Liu, and Shi]{he2024learning}
He, T., Luo, Z., Xiao, W., Zhang, C., Kitani, K., Liu, C., and Shi, G.
\newblock Learning human-to-humanoid real-time whole-body teleoperation.
\newblock \emph{arXiv preprint arXiv:2403.04436}, 2024{\natexlab{b}}.

\bibitem[Ionescu et~al.(2014)Ionescu, Papava, Olaru, and Sminchisescu]{h36m_pami}
Ionescu, C., Papava, D., Olaru, V., and Sminchisescu, C.
\newblock Human3.6m: Large scale datasets and predictive methods for 3d human sensing in natural environments.
\newblock \emph{IEEE Transactions on Pattern Analysis and Machine Intelligence}, 36\penalty0 (7):\penalty0 1325--1339, jul 2014.

\bibitem[Ji et~al.(2024)Ji, Peng, Liu, Li, Yang, Cheng, and Wang]{ji2024exbody2}
Ji, M., Peng, X., Liu, F., Li, J., Yang, G., Cheng, X., and Wang, X.
\newblock Exbody2: Advanced expressive humanoid whole-body control.
\newblock \emph{arXiv preprint arXiv:2412.13196}, 2024.

\bibitem[Jiang et~al.(2024)Jiang, Xie, Li, Yuan, Zhu, and Zhu]{jiang2024harmon}
Jiang, Z., Xie, Y., Li, J., Yuan, Y., Zhu, Y., and Zhu, Y.
\newblock Harmon: Whole-body motion generation of humanoid robots from language descriptions.
\newblock In \emph{CoRL 2024 Workshop CoRoboLearn: Advancing Learning for Human-Centered Collaborative Robots}, 2024.

\bibitem[Kim et~al.(2024)Kim, Pertsch, Karamcheti, Xiao, Balakrishna, Nair, Rafailov, Foster, Lam, Sanketi, et~al.]{kim2024openvla}
Kim, M.~J., Pertsch, K., Karamcheti, S., Xiao, T., Balakrishna, A., Nair, S., Rafailov, R., Foster, E., Lam, G., Sanketi, P., et~al.
\newblock Openvla: An open-source vision-language-action model.
\newblock \emph{arXiv preprint arXiv:2406.09246}, 2024.

\bibitem[Kuindersma et~al.(2016)Kuindersma, Deits, Fallon, Valenzuela, Dai, Permenter, Koolen, Marion, and Tedrake]{kuindersma2016optimization}
Kuindersma, S., Deits, R., Fallon, M., Valenzuela, A., Dai, H., Permenter, F., Koolen, T., Marion, P., and Tedrake, R.
\newblock Optimization-based locomotion planning, estimation, and control design for the atlas humanoid robot.
\newblock \emph{Autonomous robots}, 40:\penalty0 429--455, 2016.

\bibitem[Li et~al.(2023)Li, Ma, Kolt, Shah, and Nguyen]{li2023dynamic}
Li, J., Ma, J., Kolt, O., Shah, M., and Nguyen, Q.
\newblock Dynamic loco-manipulation on hector: Humanoid for enhanced control and open-source research.
\newblock \emph{arXiv preprint arXiv:2312.11868}, 2023.

\bibitem[Lin et~al.(2023)Lin, Zeng, Lu, Cai, Zhang, Wang, and Zhang]{lin2023motion}
Lin, J., Zeng, A., Lu, S., Cai, Y., Zhang, R., Wang, H., and Zhang, L.
\newblock Motion-x: A large-scale 3d expressive whole-body human motion dataset.
\newblock \emph{Advances in Neural Information Processing Systems}, 36:\penalty0 25268--25280, 2023.

\bibitem[Lin et~al.(2024)Lin, Yin, Ping, Molchanov, Shoeybi, and Han]{lin2024vila}
Lin, J., Yin, H., Ping, W., Molchanov, P., Shoeybi, M., and Han, S.
\newblock Vila: On pre-training for visual language models.
\newblock In \emph{Proceedings of the IEEE/CVF Conference on Computer Vision and Pattern Recognition}, pp.\  26689--26699, 2024.

\bibitem[Liu et~al.(2023)Liu, Li, Wu, and Lee]{liu2023llava}
Liu, H., Li, C., Wu, Q., and Lee, Y.~J.
\newblock Visual instruction tuning, 2023.

\bibitem[Liu et~al.(2024{\natexlab{a}})Liu, Liu, Wang, An, Li, Zhou, Yang, Zhang, Guo, and Zhang]{liurobomamba}
Liu, J., Liu, M., Wang, Z., An, P., Li, X., Zhou, K., Yang, S., Zhang, R., Guo, Y., and Zhang, S.
\newblock Robomamba: Efficient vision-language-action model for robotic reasoning and manipulation.
\newblock In \emph{The Thirty-eighth Annual Conference on Neural Information Processing Systems}, 2024{\natexlab{a}}.

\bibitem[Liu et~al.(2024{\natexlab{b}})Liu, Wu, Li, Tan, Chen, Wang, Xu, Su, and Zhu]{liu2024rdt}
Liu, S., Wu, L., Li, B., Tan, H., Chen, H., Wang, Z., Xu, K., Su, H., and Zhu, J.
\newblock Rdt-1b: a diffusion foundation model for bimanual manipulation.
\newblock \emph{arXiv preprint arXiv:2410.07864}, 2024{\natexlab{b}}.

\bibitem[Liu et~al.(2024{\natexlab{c}})Liu, Yang, Zhong, Wang, and Yi]{liu2024mimicking}
Liu, Y., Yang, B., Zhong, L., Wang, H., and Yi, L.
\newblock Mimicking-bench: A benchmark for generalizable humanoid-scene interaction learning via human mimicking.
\newblock \emph{arXiv preprint arXiv:2412.17730}, 2024{\natexlab{c}}.

\bibitem[Loper et~al.(2023)Loper, Mahmood, Romero, Pons-Moll, and Black]{loper2023smpl}
Loper, M., Mahmood, N., Romero, J., Pons-Moll, G., and Black, M.~J.
\newblock Smpl: A skinned multi-person linear model.
\newblock In \emph{Seminal Graphics Papers: Pushing the Boundaries, Volume 2}, pp.\  851--866. 2023.

\bibitem[Lu et~al.(2024)Lu, Cheng, Li, Yang, Ji, Yuan, Yang, Yi, and Wang]{lu2024mobile}
Lu, C., Cheng, X., Li, J., Yang, S., Ji, M., Yuan, C., Yang, G., Yi, S., and Wang, X.
\newblock Mobile-television: Predictive motion priors for humanoid whole-body control.
\newblock \emph{arXiv preprint arXiv:2412.07773}, 2024.

\bibitem[Luo et~al.(2023)Luo, Cao, Kitani, Xu, et~al.]{luo2023perpetual}
Luo, Z., Cao, J., Kitani, K., Xu, W., et~al.
\newblock Perpetual humanoid control for real-time simulated avatars.
\newblock In \emph{Proceedings of the IEEE/CVF International Conference on Computer Vision}, pp.\  10895--10904, 2023.

\bibitem[Mahmood et~al.(2019)Mahmood, Ghorbani, Troje, Pons-Moll, and Black]{mahmood2019amass}
Mahmood, N., Ghorbani, N., Troje, N.~F., Pons-Moll, G., and Black, M.~J.
\newblock Amass: Archive of motion capture as surface shapes.
\newblock In \emph{Proceedings of the IEEE/CVF international conference on computer vision}, pp.\  5442--5451, 2019.

\bibitem[Makoviychuk et~al.(2021)Makoviychuk, Wawrzyniak, Guo, Lu, Storey, Macklin, Hoeller, Rudin, Allshire, Handa, et~al.]{makoviychuk2021isaac}
Makoviychuk, V., Wawrzyniak, L., Guo, Y., Lu, M., Storey, K., Macklin, M., Hoeller, D., Rudin, N., Allshire, A., Handa, A., et~al.
\newblock Isaac gym: High performance gpu-based physics simulation for robot learning.
\newblock \emph{arXiv preprint arXiv:2108.10470}, 2021.

\bibitem[Mao et~al.(2024)Mao, Zhao, Song, Shi, Ye, Zhang, Geng, Malik, Guizilini, and Wang]{mao2024learning}
Mao, J., Zhao, S., Song, S., Shi, T., Ye, J., Zhang, M., Geng, H., Malik, J., Guizilini, V., and Wang, Y.
\newblock Learning from massive human videos for universal humanoid pose control.
\newblock \emph{arXiv preprint arXiv:2412.14172}, 2024.

\bibitem[Schulman et~al.(2017)Schulman, Wolski, Dhariwal, Radford, and Klimov]{schulman2017proximal}
Schulman, J., Wolski, F., Dhariwal, P., Radford, A., and Klimov, O.
\newblock Proximal policy optimization algorithms.
\newblock \emph{arXiv preprint arXiv:1707.06347}, 2017.

\bibitem[Tevet et~al.(2023)Tevet, Raab, Gordon, Shafir, Cohen-or, and Bermano]{tevet2023human}
Tevet, G., Raab, S., Gordon, B., Shafir, Y., Cohen-or, D., and Bermano, A.~H.
\newblock Human motion diffusion model.
\newblock In \emph{The Eleventh International Conference on Learning Representations}, 2023.
\newblock URL \url{https://openreview.net/forum?id=SJ1kSyO2jwu}.

\bibitem[Tong et~al.(2024)Tong, Ding, Wang, Zhang, Cui, Sun, Fan, Zhao, Zhang, Dang, et~al.]{tong2024quart}
Tong, X., Ding, P., Wang, D., Zhang, W., Cui, C., Sun, M., Fan, Y., Zhao, H., Zhang, H., Dang, Y., et~al.
\newblock Quart-online: Latency-free large multimodal language model for quadruped robot learning.
\newblock \emph{arXiv preprint arXiv:2412.15576}, 2024.

\bibitem[Van Den~Oord et~al.(2017)Van Den~Oord, Vinyals, et~al.]{van2017neural}
Van Den~Oord, A., Vinyals, O., et~al.
\newblock Neural discrete representation learning.
\newblock \emph{Advances in neural information processing systems}, 30, 2017.

\bibitem[Wang et~al.(2025)Wang, Wang, Liu, and Daniilidis]{wang2025tram}
Wang, Y., Wang, Z., Liu, L., and Daniilidis, K.
\newblock Tram: Global trajectory and motion of 3d humans from in-the-wild videos.
\newblock In \emph{European Conference on Computer Vision}, pp.\  467--487. Springer, 2025.

\bibitem[Xu et~al.(2024)Xu, Zhang, Li, Han, and Lu]{xu2024humanvla}
Xu, X., Zhang, Y., Li, Y.-L., Han, L., and Lu, C.
\newblock Humanvla: Towards vision-language directed object rearrangement by physical humanoid.
\newblock \emph{arXiv preprint arXiv:2406.19972}, 2024.

\bibitem[Zhang et~al.(2023{\natexlab{a}})Zhang, Li, and Bing]{zhang2023video}
Zhang, H., Li, X., and Bing, L.
\newblock Video-llama: An instruction-tuned audio-visual language model for video understanding.
\newblock \emph{arXiv preprint arXiv:2306.02858}, 2023{\natexlab{a}}.

\bibitem[Zhang et~al.(2023{\natexlab{b}})Zhang, Zhang, Cun, Zhang, Zhao, Lu, Shen, and Shan]{zhang2023t2m}
Zhang, J., Zhang, Y., Cun, X., Zhang, Y., Zhao, H., Lu, H., Shen, X., and Shan, Y.
\newblock Generating human motion from textual descriptions with discrete representations.
\newblock In \emph{Proceedings of the IEEE/CVF conference on computer vision and pattern recognition}, pp.\  14730--14740, 2023{\natexlab{b}}.

\end{thebibliography}
